# LLM Rationale Distillation for Spacecraft Mechanism Health Monitoring

**Hao Sun[1,2,†], Deming Li[1,2], Zhenming Zheng[1,2], Ye Xu[3,†], Fei Yang[1,2,4], Honghao Yue[1,2]**

[1] Research Center of Aerospace Mechanism and Control, Harbin Institute of Technology
[2] Laboratory of Aerospace Embodied Intelligence Technology and Equipment, Harbin Institute of Technology
[3] Wuhan Tiantuo Aerospace Intelligent Equipment Co., LTD.
[4] Suzhou Research Institute of HIT
sunnice@hit.edu.cn; xuyehit@163.com

[†]Corresponding author.

## Abstract

We study rationale distillation from large language models (LLMs) for health monitoring of a motor-driven spacecraft deployment hinge. The data are simulated life-test records with four diagnostic labels. We distill 11 LLM teachers into a Qwen2.5-3B-Instruct student with LoRA. Tests use one in-distribution split and three out-of-distribution splits that shift temperature, hardware individuals, or sensor noise. Our question is how to choose the teacher. A teacher's zero-shot score did not predict its student's score (Spearman $\rho = -0.07$ over nine teachers, $p = 0.88$), and neither did the pass rate of our rationale filter. Student scores tracked the number of rationales that passed the filter ($\rho = 0.92$). The two students that collapsed in output format came from the two teachers with the longest rationales, a post hoc pattern with two cases. Rationale-only training failed for both teachers tested. A label-only control trained on all 8000 labeled records scored 0.868 macro F1 in distribution, about 9 points above the best student. The students do add an evidence chain that quotes real telemetry values. XGBoost scored highest in distribution (0.904). Under sensor noise, the label-only control, a tuned GRU, and a Transformer were level. In this league, the number and length of usable rationales were more informative for teacher selection than the teacher's own accuracy.

## 1 Introduction

Spacecraft rely on deployable mechanisms. Solar arrays, antennas, and booms launch stowed and unfold on orbit, and a deployment failure can end a mission. Before flight, these mechanisms go through ground life tests whose telemetry carries signs of wear, obstruction, and sensor drift. Turning this telemetry into a structured diagnosis is a health monitoring task within prognostics and health management (PHM) [1, 2, 3]. Our system under test is a motor-driven deployment hinge with an end-of-travel latch (Figure 1a). A physics simulator of its life test provides 8000 training, 400 validation, and 800 in-distribution test records. Each record holds the snapshot history of one unit, with 19 telemetry indicators per snapshot. A monitoring model fills four classification heads: fault mode, severity, remaining-life bin, and maintenance action. LLMs can read such a table as text and return a diagnosis with evidence and reasoning. Hosted LLMs are too large for on-board use. Their outputs can still train a small student through rationale distillation [4, 5]. Our student is Qwen2.5-3B-Instruct [6], a 3B open-weight model fine-tuned with LoRA [7]. This leaves a practical choice open: which teacher should write the rationales? A common default is the teacher with the highest task accuracy. We test this default with 11 teachers that pass through one generation pipeline, one quality filter, and one training recipe. We also probe each teacher's own zero-shot accuracy on the same task. This lets us compare a teacher's own ability with the teachability of its outputs, measured by the score of its student.

The students are also compared with a label-only control and with 21 classical baselines on four test conditions. All scores are macro F1 over the four heads.

First, the teacher's own accuracy was a poor guide. Its zero-shot score did not predict the student's score (ρ = -0.07 over nine teachers), and the filter's pass rate did not either. Student scores tracked the number of rationales that passed the filter (ρ = 0.92). This number depended on per-teacher generation budgets as well as on the teacher, so the two effects are mixed in our league. Two students collapsed in output format, and both came from the two teachers with the longest rationales. Three teachers with similarly few but shorter rationales gave valid students. We report rationale length as a cheap warning sign that still needs a controlled test.

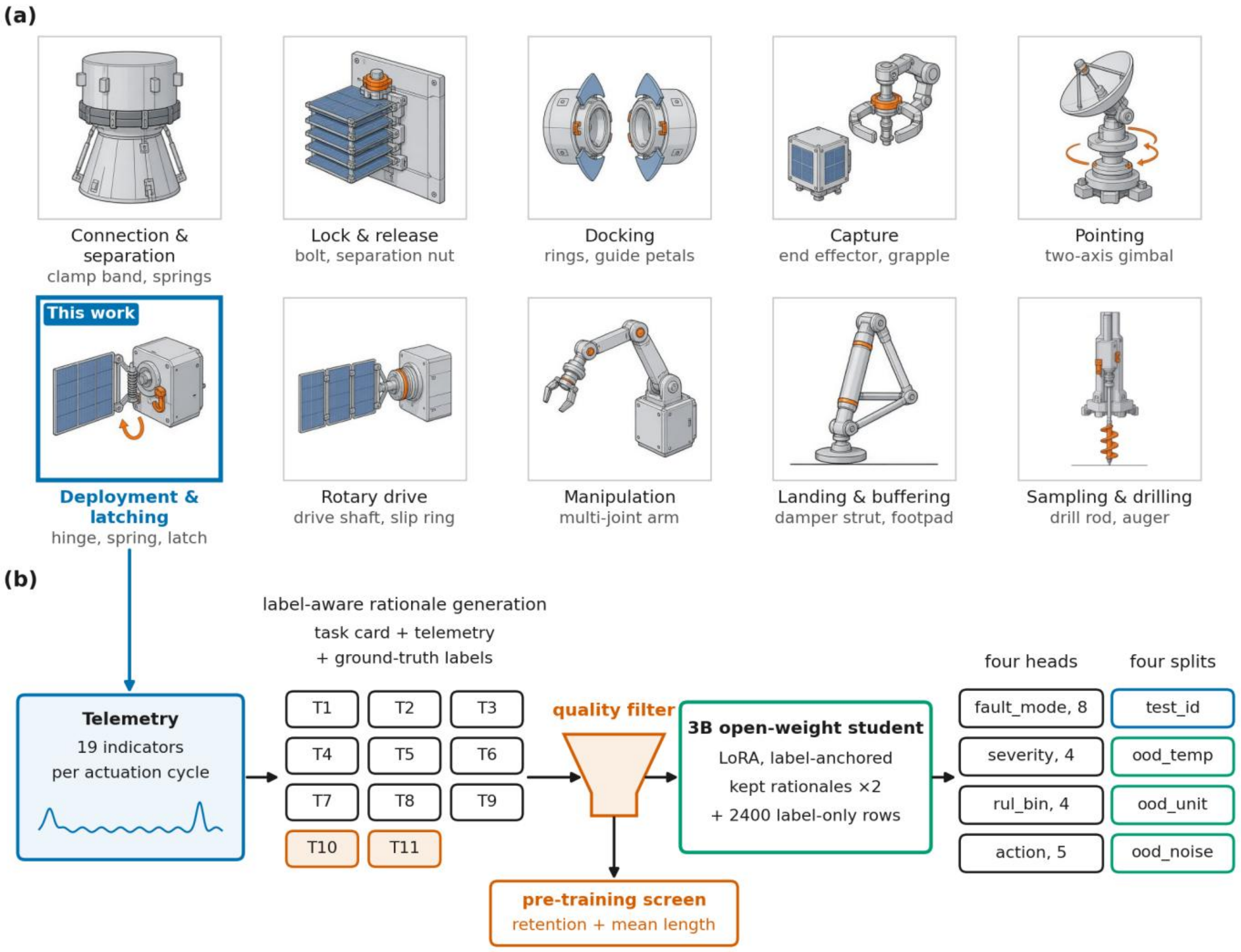

**Figure 1.** (a) Common spacecraft mechanism families; this work studies a motor-driven deployment hinge with an end-of-travel latch (19 telemetry indicators per actuation cycle). (b) Eleven teachers write rationales, a deterministic filter keeps those that pass its checks, and a Qwen2.5-3B-Instruct student with LoRA learns from the kept rationales plus label-only rows. Orange marks the two teachers whose students collapsed.

Second, distillation did not beat fine-tuning on labels alone. Rationale-only training failed for both teachers tested, and adding label-only rows restored it. A label-only control trained on all 8000 labeled records scored about 9 points above the best student in distribution. It also scored above every student on three of the four splits. The students do add an evidence chain that quotes real telemetry values.

Third, the leading model family changed with the test condition. XGBoost led in distribution, in line with earlier findings on tabular data [8]. It also led under the unit shift. Under sensor noise, the label-only control, a tuned GRU, and a Transformer were level and ahead of every tabular model.

Our contributions are as follows.

1. A teacher-comparison testbed for spacecraft mechanism health monitoring, with 11 teachers, one 3B LoRA student, a label-only control, 21 classical baselines, and four test conditions.
2. Evidence that, in this league, neither teacher zero-shot accuracy nor filter retention predicted student quality, while the number of filter-passing rationales did. Rationale length flagged the two format-collapsed students.
3. A negative result on accuracy. Rationale-only training failed, and no distilled student reached the label-only control in distribution.
4. A split-by-split comparison with tabular and sequence baselines, which shows that the leading family depends on the type of shift.

Sections 2 to 5 cover related work, the task, the method, and the setup. Section 6 reports results, Section 7 analyzes them, and Section 8 states the limitations. Appendix A defines the terms and abbreviations.

# 2 Related Work

## 2.1 Distillation into small language models

Knowledge distillation transfers ability from a large model to a smaller one [9]. Recent surveys cover distillation for LLMs [10] and model compression [11]. Rationale distillation adds teacher-written reasoning to the supervision. Students fine-tuned on teacher reasoning can exceed the prompted teacher [4], and training on labels and rationales together can match larger models with less data [5]. Small students can also be specialized toward multi-step reasoning at some cost in generic ability [12]. MiniLLM replaces the forward divergence with a reverse divergence for white-box distillation [13]. LoRA trains low-rank updates on frozen weights [7], and we use it for every student. These studies mostly fix one teacher and ask how to distill. We instead vary the teacher under one fixed recipe.

## 2.2 Language models for time series and PHM

Pretrained language models transfer to numeric sequences, either with frozen attention and feedforward layers [14] or by reprogramming series into text prototypes [15]. Reviews cover deep learning for fault diagnosis of rotating machinery [3] and transfer learning for remaining-life prediction [16]. Sensor-based monitoring with learned models also covers crew state in long missions [17] and biomedical sensing [18, 19]. LLMs now enter PHM directly [1, 2]. Closest to our domain, language models read vibration-test records of LEO constellation satellites, and deterministic gates validate their output [20]. Other work aligns vibration features with text embeddings [21], fine-tunes LLMs for fault diagnosis [22], or encodes signals as text for joint diagnosis and remaining-life prediction [23]. Domain knowledge, knowledge graphs, and maintenance text also support LLM fault diagnosis [24, 25, 26, 27]. These systems train or prompt one model in one data regime, and most target bearings and rotating machines. To our knowledge, none uses several LLMs as rationale teachers for a small student or tests separate shifts in temperature, hardware individuals, and sensor noise.

## 2.3 Tabular baselines and distribution shift

Telemetry snapshots are tabular, and tree ensembles remain strong on medium-sized tabular data [8]. Tuned deep models close part of the gap [28], while trees stay strong on small data [29]. TabLLM serializes rows as text and helps when labels are scarce [30]. A tabular foundation model predicts small tasks without per-task training [31]. Rankings can change under distribution shift. TableShift benchmarks shift in tabular data [32]. In NLP, fine-tuned small models were stronger in distribution and LLMs stronger out of distribution [33]. We therefore compare model families split by split.

### 2.4 Selecting generated training data

Careful selection can reduce the data needed for instruction tuning. A small curated set can match much larger collections [34], and keeping the part that a strong model scores highly can beat the full set [35]. Other work selects data by complexity, quality, and diversity [36], by influence on a target task [37], or with a weaker scoring model [38]. These methods score chat-style pairs, mostly with models. Our filter applies deterministic checks to diagnostic rationales, and we ask whether its pass rate predicts student quality.

## 3 Task and Data

### 3.1 System and signal

A brushed DC motor drives the hinge through a gearbox under closed-loop speed control. A torsion spring assists early in the stroke, and a latch locks the hinge at the end of travel. Deployment and release mechanisms of the same class are also built on shape memory alloy actuators [39].

The control loop shapes the signal. Extra mechanical resistance shows up mainly as extra motor current, and it lengthens deployment time only once the current limit is reached. Motor current is therefore the main carrier of health information. The spring assist can make the early-stroke current negative. Cold raises friction, grease drag, and harness stiffness for every unit, so snapshots have to be read together with temperature.

### 3.2 Records, labels, and metric

Each record follows one unit through its life test. Its earliest snapshot is the post-acceptance baseline, and later snapshots come from later cycles. Each snapshot holds 19 indicators of temperature, timing, angle, speed, motor current, voltage, latch behavior, and vibration (Appendix B, Table B1). LLMs read the snapshot history as a text table. Tabular models read the same statistics, and sequence models read the raw cycle arrays. Optical tactile skins [40] could add contact measurements in future test setups.

Each record carries four classification labels. The fault_mode head names one of eight health classes: normal, six mechanism degradations, and sensor_fault (Table B2). The severity head has 4 levels, the rul_bin head has 4 remaining-life bins, and the action head has 5 maintenance actions (Table B3). A regression target, torque_margin, is part of the output format but is not analyzed here.

Figure 2 shows how the classes occupy the telemetry space. Each fault class departs from the normal population on its own signature indicators.

The main metric is macro F1 averaged over the four classification heads. With H = {fault_mode, severity, rul_bin, action} and C(h) the class set of head h, the score is

$$MacroF1 = \frac{1}{4}\sum_{h \in H} \frac{1}{|C_h|} \sum_{c \in C_h} \frac{2TP_{h,c}}{2TP_{h,c} + FP_{h,c} + FN_{h,c}}$$

where TP(h,c), FP(h,c), and FN(h,c) count true positives, false positives, and false negatives for class c of head h. Invalid or unparseable outputs count as errors for every method.

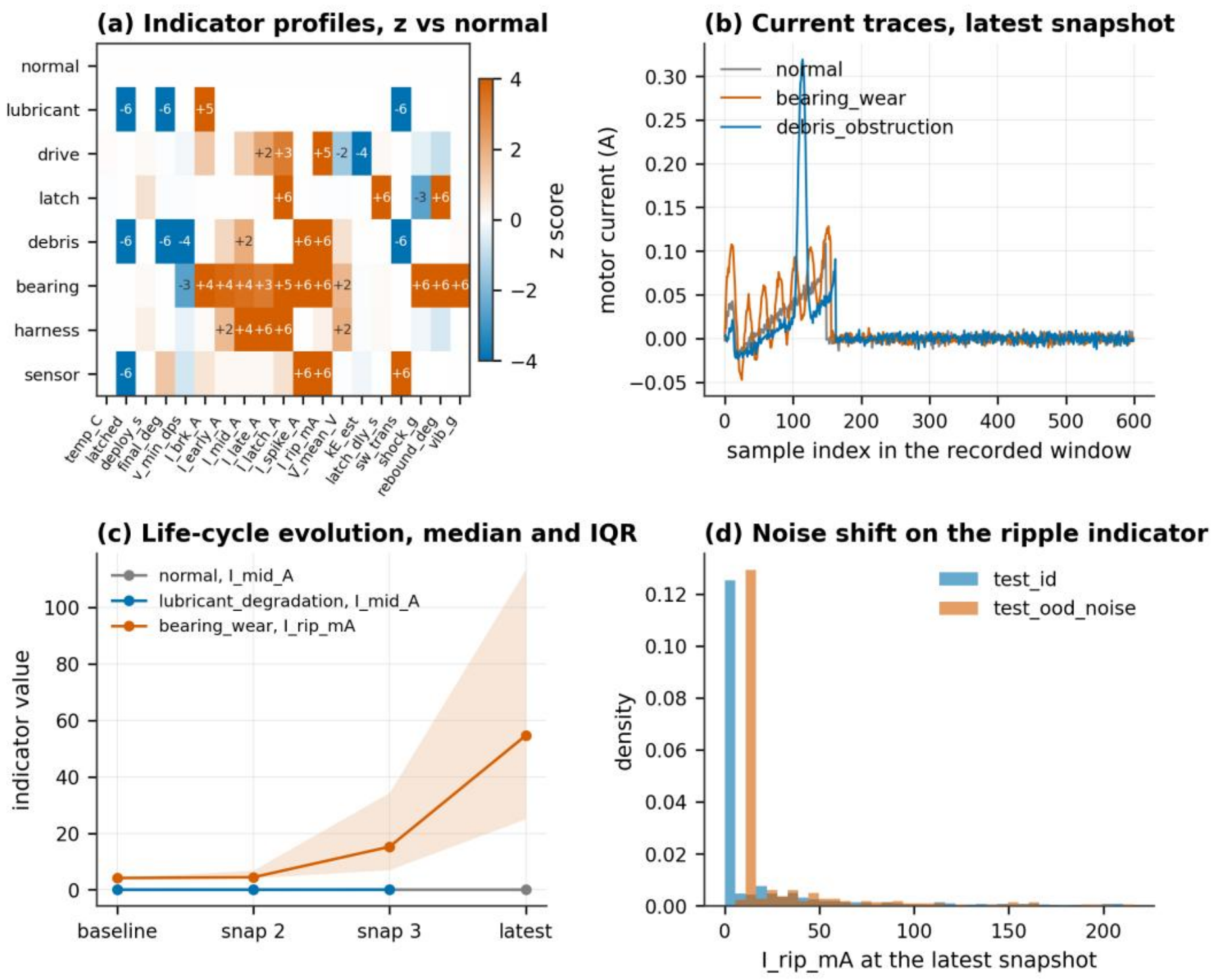

**Figure 2.** Telemetry overview, from the training split unless noted. (a) Mean of the latest snapshot per class, z-scored against the normal population per indicator and clipped at plus or minus 6. Cells with |z| ≥ 1.5 are annotated. (b) Motor current over the recorded window at the latest snapshot, for the earliest unit in train order of three classes. (c) Median and interquartile range of two signature indicators across the four life-cycle snapshots for three classes. (d) Distribution of the current-ripple indicator at the latest snapshot in test_id against test_ood_noise.

## 3.3 Splits and leakage rules

Table 1 lists the splits. The fault_mode label is balanced by construction in every split. Each out-of-distribution split shifts one factor and keeps the rest of the generating process. We write the four test conditions as test_id, ood_temp, ood_unit, and ood_noise. We use a simulator because public PHM benchmarks, such as the turbofan data, cover aero-engines rather than deployment mechanisms.

**Table 1.** The data splits. Each out-of-distribution split shifts one factor of the simulator.

| Split | Records | What shifts |
|---|---|---|
| train | 8000 | nothing, this is the reference distribution |
| val | 400 | nothing, same distribution as train |
| test_id | 800 | nothing in the generating process, only the units are new |
| test_ood_temp | 400 | the operating temperature shifts away from the training range |
| test_ood_unit | 400 | the hardware individuals shift to a different unit population |
| test_ood_noise | 400 | the sensor noise level shifts upward |

No test split is used to generate rationales or to train any model. Teachers write rationales for train records only, with the ground-truth labels in the prompt. The label-only rows also come from train. The zero-shot probe (Section 4.4) shows teachers unlabeled test_id inputs. It is an evaluation step and produces no training data.

# 4 Method

## 4.1 Rationale generation

Figure 1b shows the pipeline. We draw a class-balanced sample of train records with seed 0. Classes are interleaved, so every prefix of the sample stays balanced and a larger sample extends a smaller one. For each sampled record, the teacher receives a task card, the telemetry table, and the ground-truth labels. The task card describes the mechanism, the indicators, the health classes, and the label definitions. The teacher returns one JSON object with 2 to 4 evidence items, one reasoning paragraph, and the labels copied unchanged. Each evidence item quotes an indicator, a cycle, and a value from the table. The prompt asks the teacher to justify the given labels as an engineer reading the data, without saying that the labels were given. The rationales are thus label-conditioned explanations rather than independent diagnoses.

## 4.2 Quality filter and retention rate

A rationale enters training only if it passes four deterministic checks.

- **Schema check.** The output parses as one JSON object with 2 to 4 non-empty evidence strings and a non-empty reasoning paragraph.
- **Evidence faithfulness.** Every number quoted in the evidence appears in the telemetry table of the same record.
- **Label consistency.** The four class labels equal the ground truth, and the torque margin matches the reference value within a fixed tolerance.
- **Leak check.** A pattern match rejects phrases that refer to ground truth or to provided labels.

The retention rate is the share of completed generations that pass all four checks. Retention and the mean length of the kept rationales are both known before any student is trained. Section 7.2 tests them as pre-training screens (Figure 1b).

## 4.3 Student recipes and the label-only control

All students share the base model Qwen2.5-3B-Instruct [6] and one set of LoRA hyperparameters (Appendix D). The v2 recipe, used for all main results, mixes two kinds of rows. Each kept rationale appears twice, with evidence, reasoning, and labels as the target. Added to these are 2400 class-balanced label-only rows from the head of the training file, with the labels alone as the target. They are added after the validation split is carved out, so validation stays rationale-only. The v1 recipe is an ablation that trains on the kept rationales only. The label-only control fine-tunes the same base model on all 8000 train records with labels alone as the target. The three recipes therefore differ in training volume as well as in content.

## 4.4 Zero-shot teacher probe

The probe measures a teacher's own diagnostic ability. The teacher receives the same task card and telemetry table without labels. It answers a prefix-balanced sample of up to 150 test_id records drawn with seed 0. Its outputs go through the student scorer, and invalid outputs count as errors.

# 5 Experimental Setup

## 5.1 Teachers

The 11 teachers mix general chat models and reasoning-oriented models from seven providers. DeepSeek supplies T1 (DeepSeek-V4.1-Flash), T2 (DeepSeek-V3), and T5 (DeepSeek-R1). The Qwen team at Alibaba supplies T3 (Qwen2.5-72B-Instruct), T4 (Qwen3-32B), and T7 (Qwen3.5-27B). The other teachers are T6 (Seed-OSS-36B-Instruct, ByteDance Seed), T8 (Ling-flash-2.0, inclusionAI), T9 (Kimi-K2.6, Moonshot AI), T10 (Hy4-preview, Tencent), and T11 (GLM-5.3, Zhipu AI). All teachers are called through the SiliconFlow API, and Table D1 lists the model identifiers. The labels T1 to T11 follow the test_id scores of the students. Every teacher uses the same prompts, sampling plan, and filter. Generation targets were set per teacher for cost control, so the number of kept rationales ranges from 111 to 2044. Nine teachers have complete probes. The API budget ran out after 5 of 40 records of the T11 probe, and T5 has no probe result.

## 5.2 Students, control, and baselines

Each distilled student is a single training run. The label-only control reports the mean of 2 seeds, with a test_id standard deviation of 0.005. Sixteen tabular baselines range from logistic regression to gradient-boosted trees such as XGBoost. Each runs with default hyperparameters and with Optuna tuning on the validation split. Five sequence architectures (RNN, LSTM, GRU, Transformer, and 1D CNN) run in seven configurations on the raw cycle arrays. Every baseline configuration reports the mean over 3 seeds. Appendix C lists all baseline results.

## 5.3 Protocol and statistics

All methods share the splits and the scorer. Rank correlations use Spearman's ρ with permutation p-values. Paired bootstraps resample test rows 10000 times and use the stored predictions of single runs (Section 7.4). Appendix D gives implementation, cost, and budget details.

# 6 Results

All scores in this section are macro F1 over the four classification heads.

## 6.1 Distillation league

Table 2 lists the league with the model name of each teacher. Probe is the teacher's own zero-shot score. Kept, retention, and mean chars describe the filtered rationales and are known before training. Mean chars is the mean length of the reasoning field. Valid rate is the share of student outputs that parse with every field present.

**Table 2.** The distillation league, sorted by student test_id score. Probe is the teacher's zero-shot score on 100 to 150 test_id records. T5 has no probe. The T11 probe stopped after 5 of 40 records, so its value in parentheses is not comparable. Students are single runs.

| Teacher | Name | Probe | Kept | Retention | Mean chars | test_id | ood_temp | ood_unit | ood_noise | Valid rate |
|---|---|---|---|---|---|---|---|---|---|---|
| T1 | DeepSeek-V4.1-Flash | 0.503 | 2044 | 0.82 | 672 | 0.780 | 0.745 | 0.701 | 0.678 | 0.999 |
| T2 | DeepSeek-V3 | 0.383 | 687 | 0.86 | 436 | 0.773 | 0.733 | 0.699 | 0.678 | 0.996 |

| Teacher | Name | Probe | Kept | Retention | Mean chars | test_id | ood_temp | ood_unit | ood_noise | Valid rate |
|---|---|---|---|---|---|---|---|---|---|---|
| T3 | Qwen2.5-72B-Instruct | 0.422 | 778 | 0.97 | 725 | 0.767 | 0.772 | 0.712 | 0.641 | 1.000 |
| T4 | Qwen3-32B | 0.375 | 584 | 0.73 | 641 | 0.673 | 0.622 | 0.604 | 0.575 | 0.988 |
| T5 | DeepSeek-R1 | n/a | 226 | 0.49 | 403 | 0.670 | 0.650 | 0.625 | 0.602 | 0.990 |
| T6 | Seed-OSS-36B-Instruct | 0.405 | 288 | 0.72 | 523 | 0.641 | 0.604 | 0.584 | 0.580 | 0.985 |
| T7 | Qwen3.5-27B | 0.582 | 122 | 0.99 | 579 | 0.589 | 0.569 | 0.533 | 0.504 | 0.943 |
| T8 | Ling-flash-2.0 | 0.202 | 178 | 0.22 | 680 | 0.536 | 0.519 | 0.480 | 0.450 | 0.993 |
| T9 | Kimi-K2.6 | 0.459 | 146 | 0.92 | 709 | 0.515 | 0.482 | 0.489 | 0.481 | 0.951 |
| T10 | Hy4-preview | 0.489 | 169 | 0.88 | 1215 | 0.390 | 0.401 | 0.400 | 0.357 | 0.621 |
| T11 | GLM-5.3 | (0.669) | 111 | 0.35 | 1327 | 0.285 | 0.275 | 0.276 | 0.254 | 0.436 |

The T1 student scored highest on test_id (0.780). The T3 student scored highest on ood_temp (0.772) and ood_unit (0.712), and the T1 and T2 students tied on ood_noise (0.678). The T10 and T11 students trailed on every split. Their valid rates of 0.621 and 0.436 point to output-format failure, and we call them collapsed. Figure 3 summarizes the league.

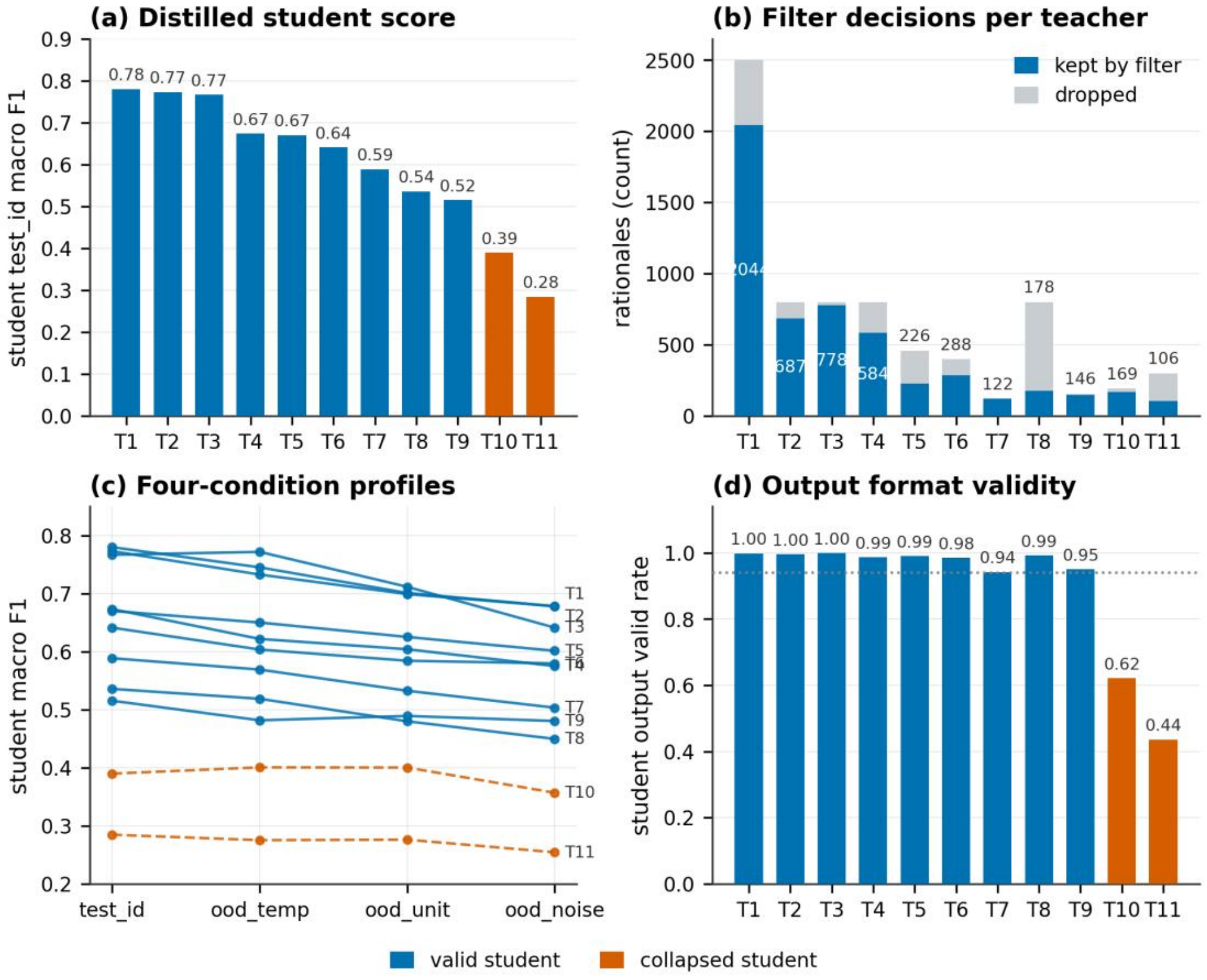

**Figure 3.** The distillation league at a glance. (a) Student test_id score per teacher. (b) Filter decisions per teacher, stacked kept and dropped counts. (c) Student scores across the four conditions, with collapsed students dashed. (d) Student output valid rate with the 0.94 line. Collapsed students are orange throughout. Students are single runs.

## 6.2 Recipes and the label-only control

Table 3 compares the recipes. Without label rows, the T3 student fell from 0.767 to 0.375 on test_id, and the T4 student fell from 0.673 to 0.271. Figure 4 shows the drop on all four splits. The v2 recipe also has several times more training rows than v1, so this ablation does not separate label anchoring from data volume.

**Table 3.** Training recipes for the same 3B student. The v1 and v2 rows of a teacher use the same kept rationales. Students are single runs, and the control is the mean of 2 seeds.

| Model | Training rows | test_id | ood_temp | ood_unit | ood_noise |
|---|---|---|---|---|---|
| T3 student, v1 | 778 rationales | 0.375 | 0.362 | 0.367 | 0.367 |
| T3 student, v2 | 778 rationales ×2 + 2400 labels | 0.767 | 0.772 | 0.712 | 0.641 |
| T4 student, v1 | 584 rationales | 0.271 | 0.285 | 0.266 | 0.239 |
| T4 student, v2 | 584 rationales ×2 + 2400 labels | 0.673 | 0.622 | 0.604 | 0.575 |
| Label-only control | 8000 labels | 0.868 | 0.840 | 0.705 | 0.752 |

The label-only control reached 0.868 on test_id, about 9 points above the best student. It scored above every distilled student on test_id, ood_temp, and ood_noise. On ood_unit the T3 student reached 0.712 against 0.705 for the control. This margin of 0.007 comes from a single-seed student, so we do not read it as a reversal. The control trains on all 8000 labeled records, while each student sees 2400 label rows plus its rationales. Part of the gap may therefore come from data volume.

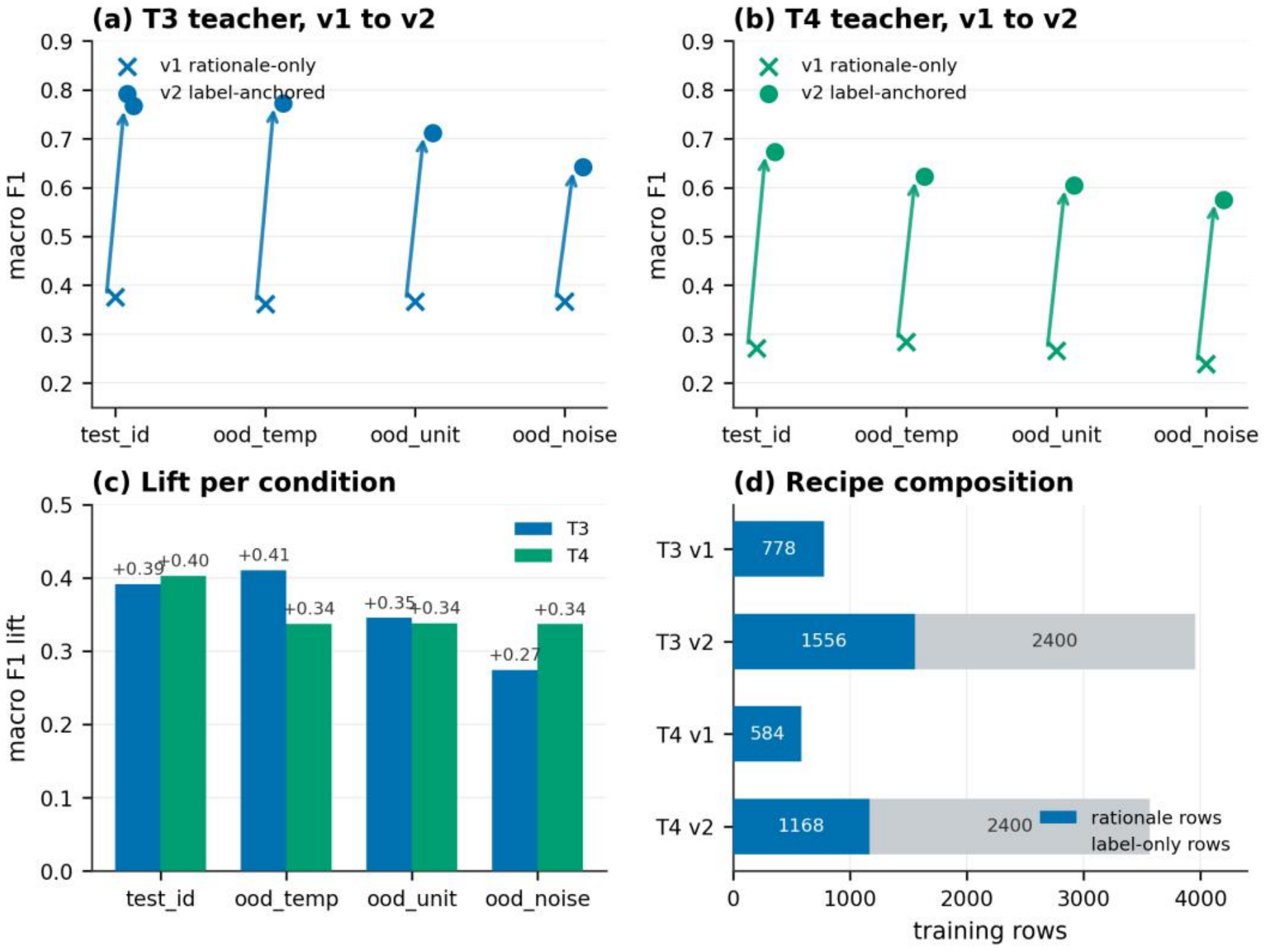

**Figure 4.** The v1 rationale-only ablation against the v2 label-anchored recipe. (a) and (b) paired scores for T3 and T4 on the four conditions, arrows from v1 to v2 with identical kept rationales. (c) The lift per condition for both teachers. (d) Training row composition of the two recipes.

### 6.3 Per-head and per-class scores

Figures 5 and 6 break the scores down by head and by fault class. On test_id, the normal class has the lowest fault_mode F1 for the label-only control and the six strongest students. For the students, one reading is that a healthy unit offers no fault signature to cite. Under sensor noise, the fault_mode F1 of XGBoost drops more than that of the LLM-based rows. Its normal-class F1 falls to 0.25, while the label-only control holds 0.72.

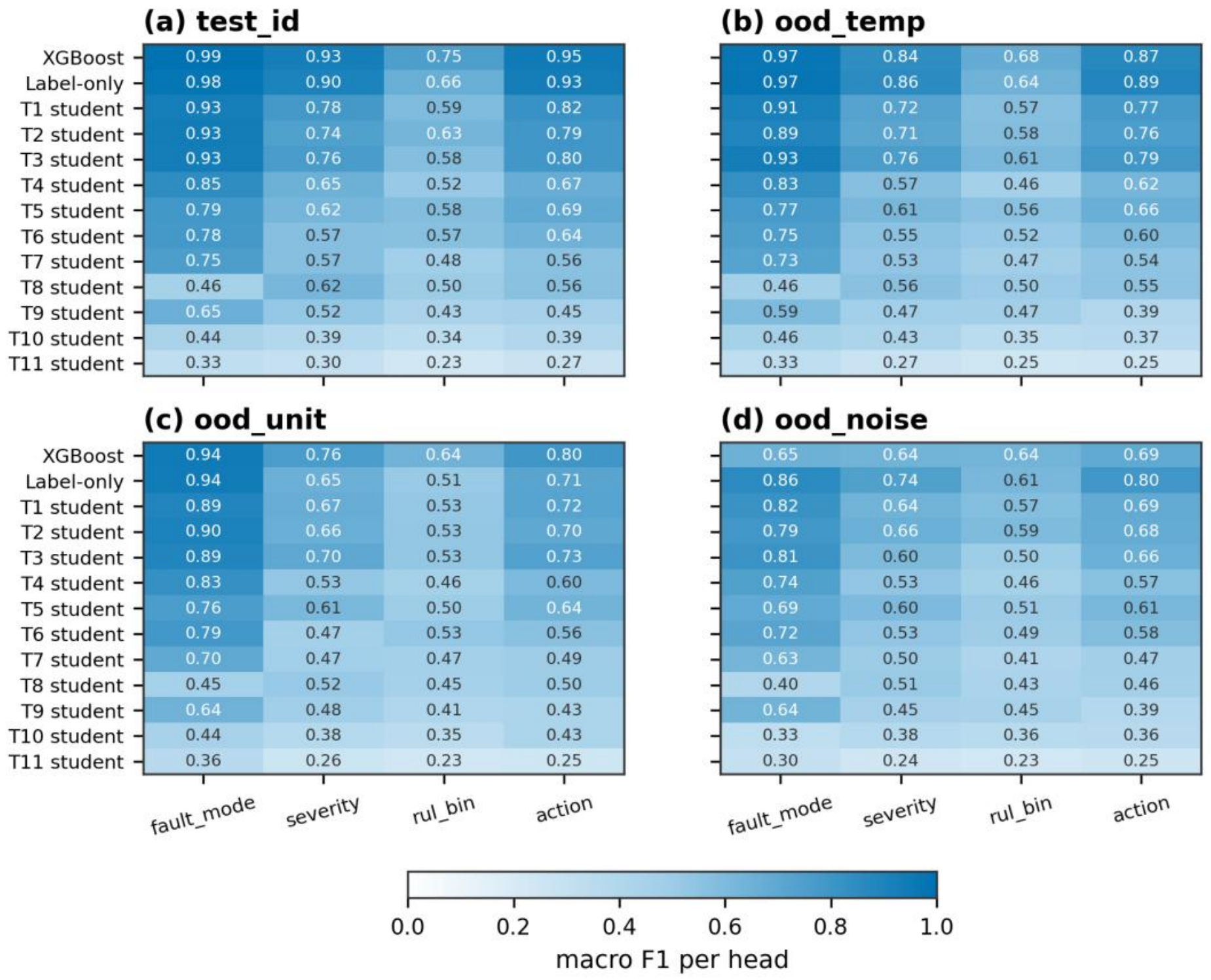

**Figure 5.** Per-head macro F1 on the four conditions. Rows are XGBoost computed from stored predictions, the label-only control (mean over 2 seeds), and the 11 distilled students (single runs). Under sensor noise the fault_mode F1 of XGBoost drops more than that of the LLM-based rows.

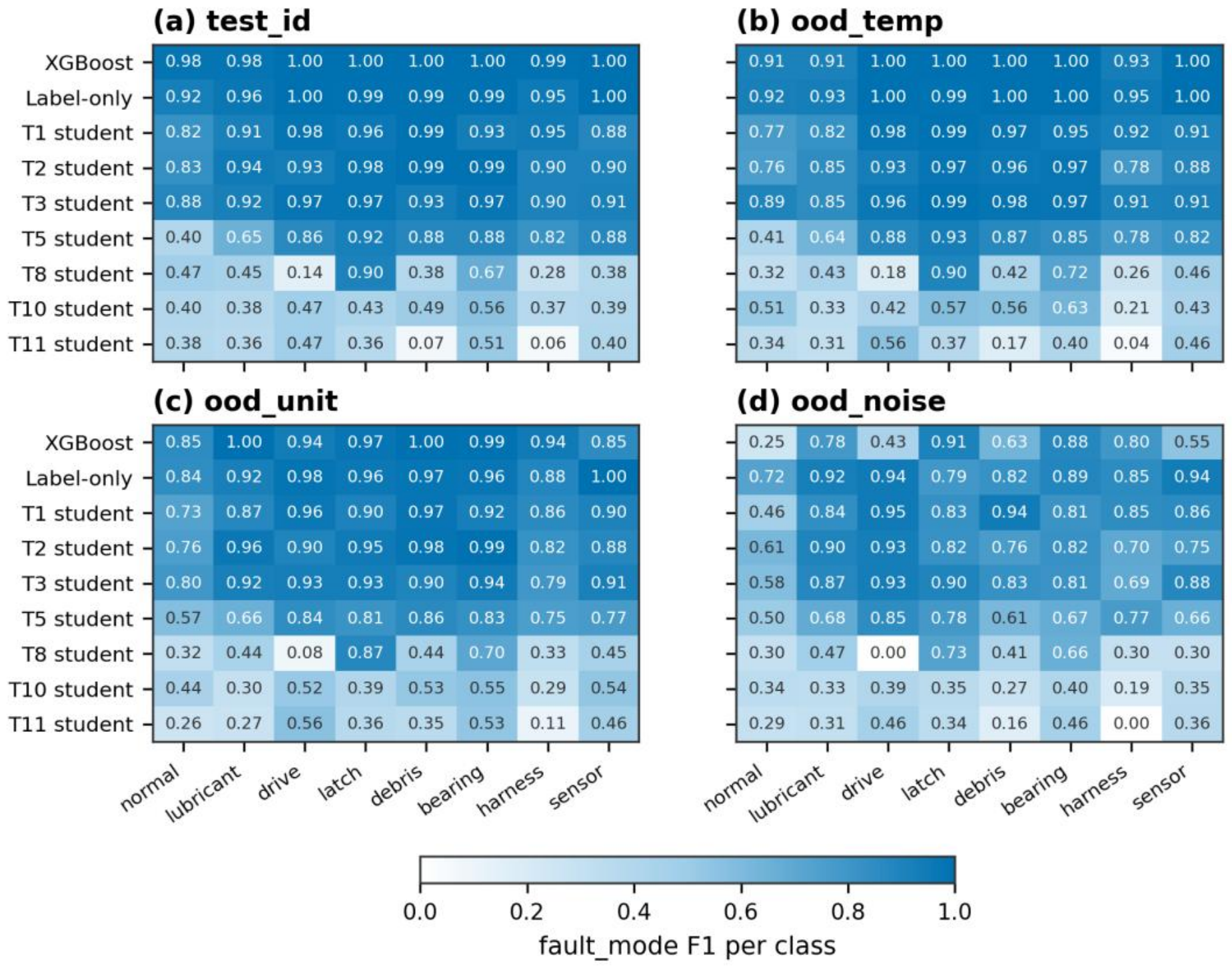

**Figure 6.** Per-class fault_mode F1 on the four conditions. Rows are XGBoost computed from stored predictions, the label-only control (mean over 2 seeds), and seven distilled students (single runs). On test_id the normal class is the weakest class for the control and the stronger students. Under sensor noise it is the class where XGBoost degrades most.

# 7 Analysis

## 7.1 Teacher ability did not predict student quality

The probe and the league rank the teachers differently (Figure 7a). Over the nine teachers with complete probes, the Spearman correlation between probe score and student test_id score is -0.07, with an exact two-sided permutation p of 0.88. Selecting by probe score would pick T7, T1, and T10 as the top three. Their students rank 7th, 1st, and 10th of 11.

Eight of the nine students scored above their teacher's probe. The reverse case is T10 (probe 0.489, student 0.390), and T7 is close (0.582 against 0.589). This comparison is not like-for-like. Each student also trains on 2400 labeled rows, and the probe covers 100 to 150 test_id records rather than all 800.

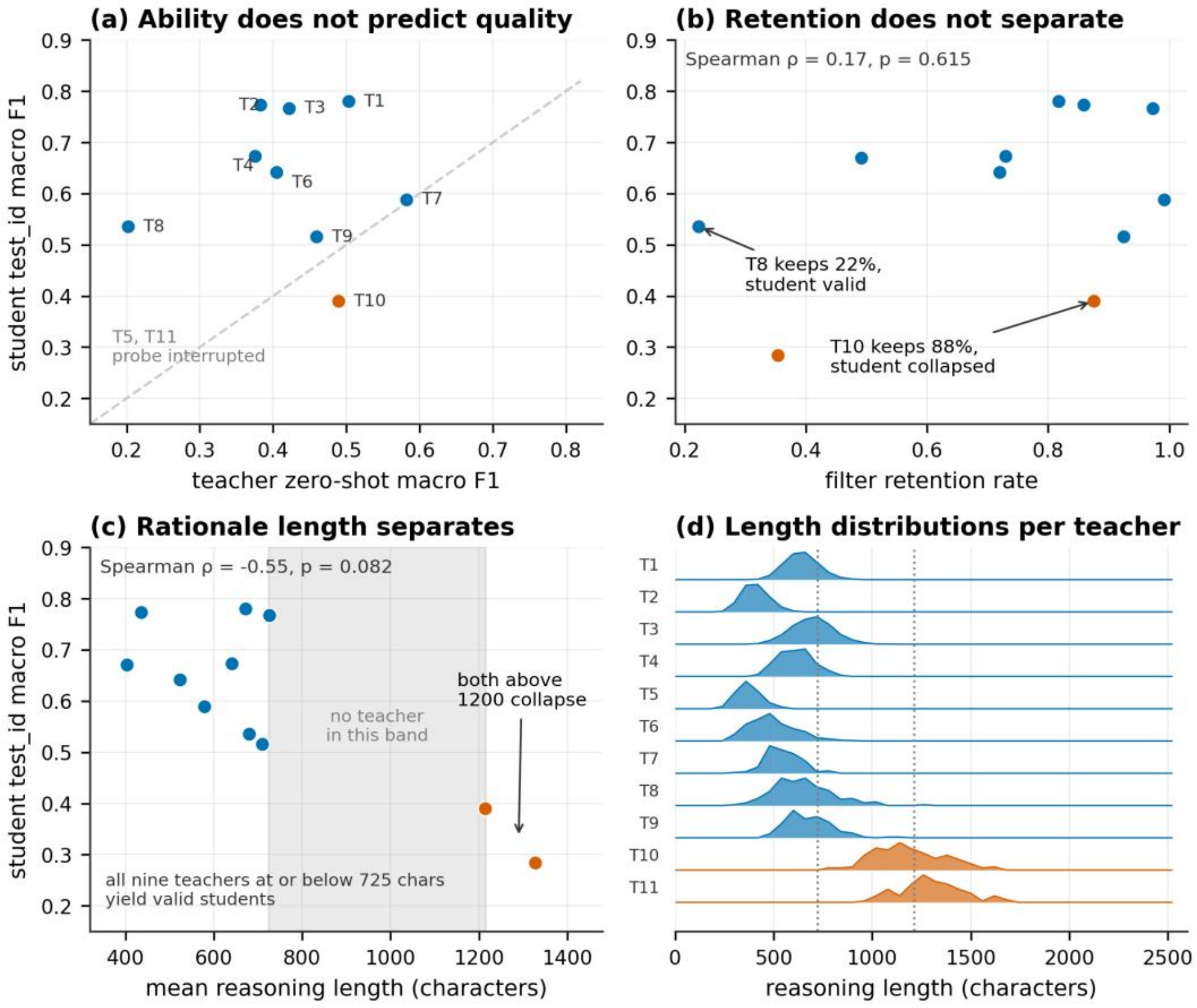

**Figure 7.** Teacher-side screens against the student test_id score. (a) Teacher zero-shot probe score for the nine teachers with complete probes (100 to 150 records each, same scorer as the students). (b) Filter retention rate (Spearman ρ = 0.17, exact permutation p = 0.615, n = 11). T8 keeps 22 percent and yields a valid student, while T10 keeps 88 percent and collapses. Retention counts completed generations only, since three teachers stopped early on budget. (c) Mean reasoning length. All nine teachers at or below 725 characters yield valid students, and both teachers above 1200 characters collapse. (d) Reasoning length distribution per teacher, rows ordered by student score, dotted lines at 725 and 1215 characters.

## 7.2 Retention, rationale count, and rationale length

Retention did not predict student quality either (Figure 7b). Its Spearman correlation with the student test_id score over the 11 teachers is 0.17 (exact permutation p = 0.615). T8 kept 22 percent of its rationales, and its student still parsed 99 percent of its outputs. T10 kept 88 percent, and its student collapsed.

The number of kept rationales tracked student quality closely. Its Spearman correlation with the student test_id score is 0.92 (Monte Carlo permutation p about 0.0002). The four teachers with the largest numbers of kept rationales gave the four best students. This number depends on the per-teacher generation target as well as on retention. Teacher identity and data volume are therefore confounded in the league.

Count alone does not explain the two collapsed students. T7, T8, and T9 kept 122 to 178 rationales, close to the 169 of T10 and the 111 of T11. The students of T7, T8, and T9 stayed valid (valid rates 0.943 to 0.993), while those of T10 and T11 did not (0.621 and 0.436). Within this low-count group, length differs. T7, T8, and T9 average 579 to 709 characters of reasoning, against 1215 for T10 and 1327 for T11 (Figure 7c). Figure 7d shows the full distributions. The two verbose teachers sit well to the right, and the valid teachers stay mostly below the 1215 line. Across the league, all nine teachers at or below 725 characters gave valid students, and no teacher falls between 725 and 1215.

This evidence is limited. The rank correlation between length and student score is weak ($\rho = -0.55$, $p = 0.082$). The separation rests on two collapsed cases, and we noticed the empty band after the runs. We therefore treat it as a hypothesis. One possible mechanism is that the student imitates long outputs and then fails to produce parseable JSON. The filter does not check length, so long rationales with real quoted values pass it. Testing the cause needs a controlled study that varies length at a fixed rationale count. Mean length can be measured on a small batch before any training, but we have not tested how small the batch can be.

### 7.3 What the rationales add

Since the label-only control scored higher, the value of distillation in this study lies in what each prediction carries. Each student prediction comes with evidence items and reasoning. On test_id, the evidence faithfulness of every student lies between 0.996 and 1.000. The collapsed T10 and T11 students score 1.000 on their parseable outputs, so their failure is format rather than fabricated evidence. This check confirms only that the quoted values exist in the input table. It does not show that the reasoning is correct or that the cited values drove the prediction. The rationales are also label-conditioned (Section 4.1), and we did not review student reasoning by hand.

### 7.4 Model families across test conditions

Table 4 lists selected configurations from each family, and Figure 8 places all 35 configurations on the four conditions. No family led on every split.

- **test_id.** XGBoost reached 0.904 with default and tuned settings, and other boosted ensembles followed closely. The label-only control reached 0.868.
- **ood_temp.** The label-only control (0.840), default HistGB (0.840), and XGBoost (0.836) were level.
- **ood_unit.** Default XGBoost held 0.787, above every LLM-based result (best 0.712). From test_id to ood_unit, the label-only control lost 0.163 and XGBoost lost 0.117. Six tabular models, including the MLP and the linear models, fell below 0.10 with both settings.
- **ood_noise.** The label-only control (0.752), the tuned GRU (0.745, seed standard deviation 0.040), and the default Transformer (0.737) were level. The best tabular score was 0.687 (default GBDT), and the best student reached 0.678.

**Table 4.** Selected configurations from each model family. Tabular and sequence rows are means over 3 seeds, the control is the mean of 2 seeds, and students are single runs. Appendix C gives the full baseline tables.

| Family | Configuration | test_id | ood_temp | ood_unit | ood_noise |
|---|---|---|---|---|---|
| Distilled student | T1, v2 | 0.780 | 0.745 | 0.701 | 0.678 |
| Distilled student | T3, v2 | 0.767 | 0.772 | 0.712 | 0.641 |
| LLM, labels only | Label-only control | 0.868 | 0.840 | 0.705 | 0.752 |
| Tabular | XGBoost, default | 0.904 | 0.836 | 0.787 | 0.642 |
| Tabular | XGBoost, tuned | 0.904 | 0.836 | 0.773 | 0.681 |
| Tabular | HistGB, default | 0.899 | 0.840 | 0.766 | 0.594 |
| Tabular | GBDT, default | 0.881 | 0.772 | 0.517 | 0.687 |
| Sequence | GRU, tuned | 0.803 | 0.746 | 0.627 | 0.745 |
| Sequence | Transformer, default | 0.770 | 0.738 | 0.506 | 0.737 |

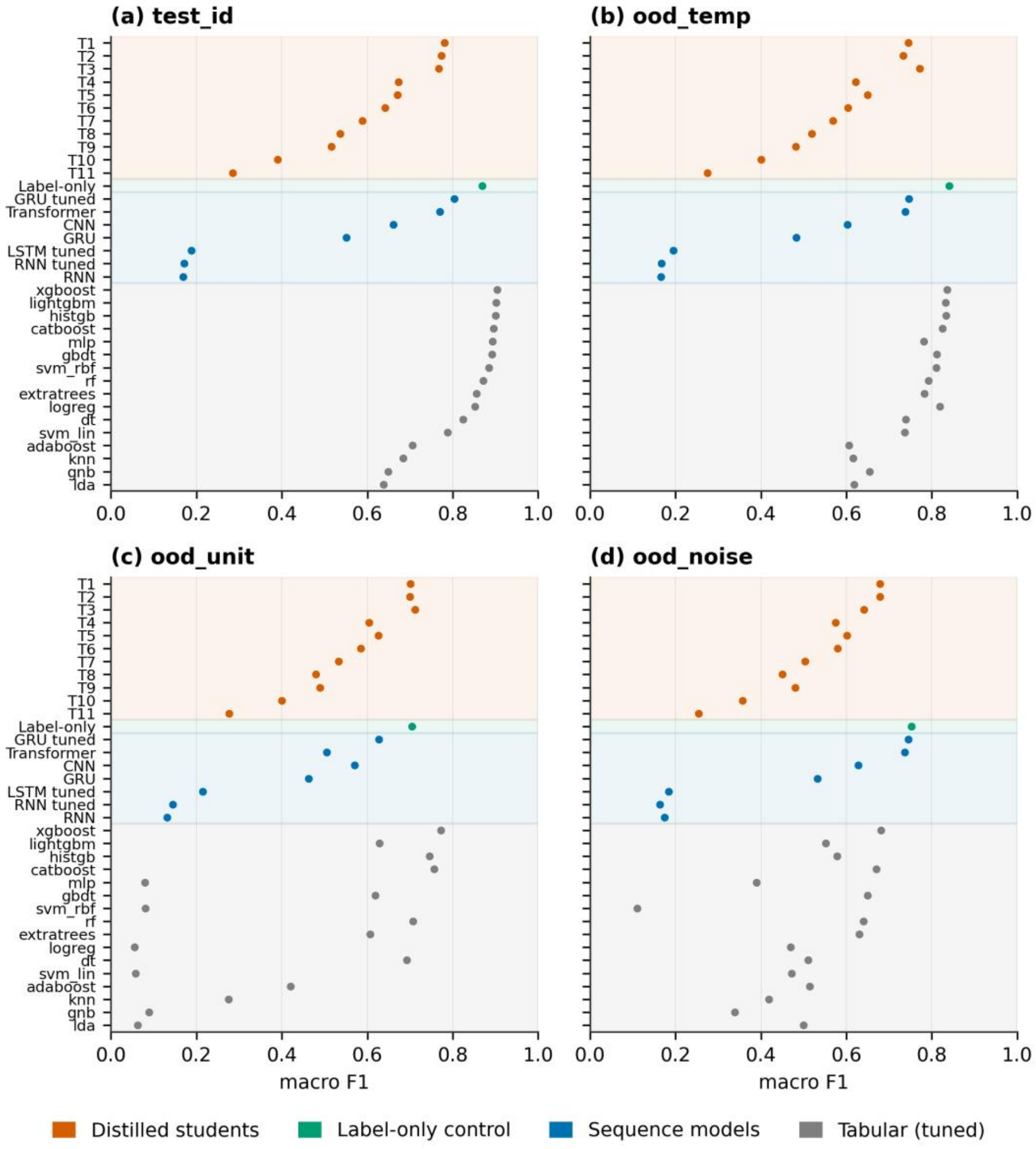

**Figure 8.** Macro F1 of every evaluated configuration on the four conditions. From top to bottom the bands hold the 11 distilled students, the label-only control, the 7 sequence configurations, and the 16 tuned tabular baselines. Tabular and sequence rows are means over 3 seeds, the label-only control is the mean of 2 seeds, and students are single runs. Tree ensembles lead on test_id and are level with the label-only control on ood_temp. The unit shift scatters the classical families, and several tabular models fall below 0.10 there. Under sensor noise the label-only control, the tuned GRU, and the default Transformer are level.

The table means mix runs and seeds, so the paired tests use the stored predictions of two single runs. These are one label-only run (seed 0) and one default XGBoost run. On ood_noise, they scored 0.732 and 0.653. The gap of 0.079 has a 95 percent bootstrap interval of [0.042, 0.114]. On test_id, they scored 0.864 and 0.904, a gap of -0.040 with an interval of [-0.056, -0.025]. Both p-values are below 0.001. These tests cover row sampling within one run each and do not cover seed variance.

Figure 9 shows where the ood_noise difference comes from. Under sensor noise, every method loses normal units to false alarms, but XGBoost also loses fault classes. It reads drive_degradation as sensor_fault in 58 percent of cases. The label-only control keeps every fault class at 0.68 or above on the diagonal, and the T1 student keeps them at 0.78 or above. For this task, XGBoost was the stronger choice in distribution and under the unit shift. Under sensor noise, the label-only control, the tuned GRU, and the default Transformer were ahead of every tabular model.

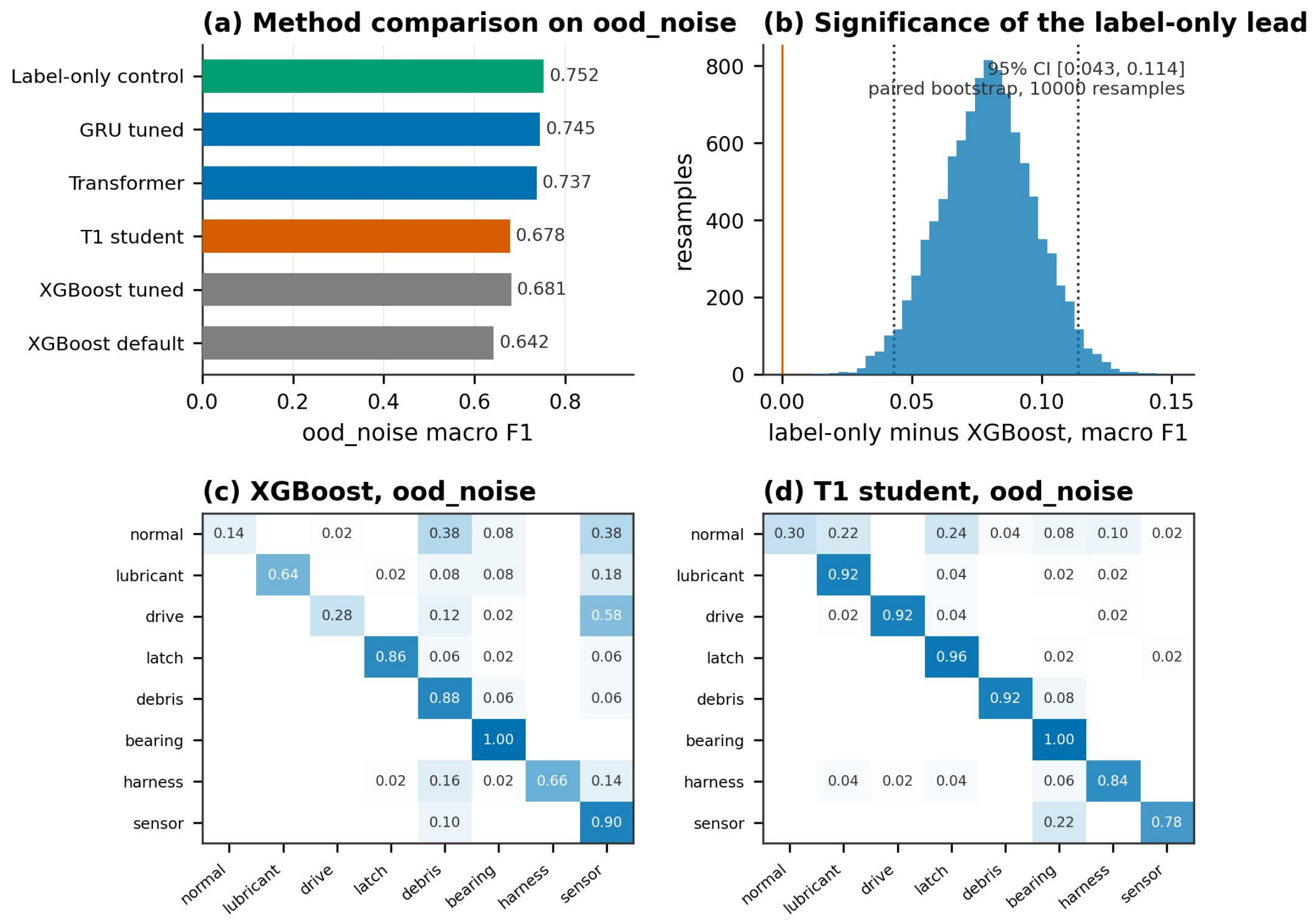

**Figure 9.** The ood_noise split in detail. (a) Macro F1 of selected methods. Sequence and tree numbers are means over 3 seeds, the label-only control is the mean of 2 seeds, and the distilled student is a single run. (b) Paired bootstrap of the gap between one label-only run (seed 0) and one stored default XGBoost run, 10000 resamples over the 400 rows, with the 95 percent interval marked. (c) and (d) Row-normalized fault_mode confusion of XGBoost (stored default run) and the T1 student (single run). Under noise the tree model loses normal units to false alarms and confuses drive_degradation with sensor_fault, while the student keeps the fault classes on the diagonal.

# 8 Limitations

- **Probe coverage.** Nine teachers have complete probes of 100 to 150 records. The T11 probe stopped after 5 of 40 records, and its score of 0.669 is not comparable. T5 has no probe. Eight teachers ran with thinking disabled, while T8 and T11 ran in default mode.

- **Confounds in the league.** Students differ in the teacher and in the number of kept rationales, which our budgets set per teacher. The length pattern rests on two collapsed cases noticed after the runs, and it may shift with the task, the prompt, or the teacher pool.
- **Single runs.** Each distilled student is one training run, so seed variance is unmeasured. The paired bootstraps also compare single runs.
- **Data volume.** The label-only control trains on more labeled rows than any student, and v2 trains on several times more rows than v1. Our recipe comparisons therefore do not separate content from volume. We tested one rationale format and one mixing recipe.
- **Scope.** All students share one 3B base model, and the teachers come from Chinese providers only. All data come from one simulator, so the sim-to-real gap is untested. On-board deployment is also untested, and reconfigurable in-memory computing hardware such as [41] could be studied as a flight-side platform.
- **Evidence quality.** Faithfulness is a numeric check, and no expert reviewed the reasoning. The torque_margin regression head is not analyzed.

## 9 Conclusion

We asked how to choose a teacher when distilling LLM rationales into a small student for spacecraft mechanism health monitoring. Across 11 teachers, neither the teacher's zero-shot score nor the filter's pass rate predicted student quality. Student quality tracked the number of filter-passing rationales, and the two format-collapsed students came from the two teachers with the longest rationales. Rationale-only training failed. A label-only control trained on more labeled data outscored the best student by about 9 points in distribution. The value of the students therefore lies in an auditable evidence trail. Across model families, XGBoost led in distribution and under the unit shift. Under sensor noise, the label-only control, a tuned GRU, and a Transformer were level. For teacher selection, we suggest generating a small batch first and checking its yield and mean length, rather than ranking teachers by their own accuracy. A controlled study of length at a fixed rationale count is the next step.

## Reproducibility

The simulator configuration, generated splits, prompts, quality filter, and training configurations are available from the corresponding author upon reasonable request. Teacher rationales were generated through the SiliconFlow API in November 2026.

# Appendix A. Nomenclature and Abbreviations

Table A1 defines the terms, symbols, and abbreviations used in the paper.

**Table A1.** Terms, symbols, and abbreviations.

| Term | Meaning |
|---|---|
| PHM | Prognostics and health management |
| LLM | Large language model |
| LoRA | Low-rank adaptation, which trains low-rank updates on frozen pretrained weights [7] |
| OOD | Out of distribution |
| RUL | Remaining useful life |
| Record | The life-test snapshot history of one unit, with its labels |
| Snapshot | The 19 telemetry indicators recorded at one actuation cycle |
| test_id | In-distribution test split of 800 records from new units |
| ood_temp, ood_unit, ood_noise | Test splits of 400 records each, with shifted temperature, hardware individuals, or sensor noise |
| Head | One classification output: fault_mode, severity, rul_bin, or action (Table B3) |
| torque_margin | Regression target, the torque margin at the -20 C reference case; not analyzed here |
| Macro F1 | Unweighted mean of per-class F1 within a head, averaged over the four heads; invalid outputs count as errors |
| Teacher, T1 to T11 | Hosted LLM that writes rationales, numbered by the test_id score of its student; names in Table 2 and Table D1 |
| Student | Qwen2.5-3B-Instruct fine-tuned with LoRA on the kept rationales of one teacher |
| Rationale | Teacher output with 2 to 4 evidence items, one reasoning paragraph, and the labels |
| Kept rationale | A rationale that passes all four filter checks |
| Retention rate | Kept rationales divided by completed generations |
| Mean chars | Mean length in characters of the reasoning field over the kept rationales |
| Valid rate | Share of student outputs that parse with all labels and the torque margin present |
| Evidence faithfulness | Share of the numbers quoted in evidence items that appear in the input telemetry table |
| Collapsed student | A student whose valid rate falls far below the rest of the league (T10 and T11) |
| Label anchoring | Mixing label-only rows into the rationale training set |
| v1 recipe | Kept rationales only |
| v2 recipe | Each kept rationale twice plus 2400 label-only rows; used for all main results |
| Label-only control | The same base model fine-tuned on all 8000 train records with labels only |
| Zero-shot probe | A teacher answers unlabeled test_id records and is scored like the students |
| Teachability | The test_id score of the student trained on the kept rationales of a teacher under v2; it mixes output style and rationale count |
| ρ | Spearman rank correlation |
| GBDT, HistGB | Gradient-boosted decision trees; histogram-based gradient boosting |

| Term | Meaning |
|---|---|
| XGBoost, LightGBM, CatBoost | Gradient-boosting libraries |
| RF, DT, ExtraTrees | Random forest; decision tree; extremely randomized trees |
| SVM, kNN, LDA, GNB, MLP | Support vector machine; k-nearest neighbors; linear discriminant analysis; Gaussian naive Bayes; multilayer perceptron |
| RNN, LSTM, GRU, CNN | Recurrent network; long short-term memory; gated recurrent unit; convolutional network |
| CNY | Chinese yuan |

## Appendix B. Telemetry Indicators and Health Classes

Table B1 lists the 19 indicators recorded at each actuation cycle. Table B2 gives the typical signature of each health class. Table B3 defines the labels.

**Table B1.** The 19 telemetry indicators recorded for each actuation cycle of the hinge life test.

| Indicator | Recorded quantity |
|---|---|
| temp_C | mechanism temperature |
| latched | latch microswitch state at the end of travel |
| deploy_s | time from motion start to switch closure |
| final_deg | potentiometer angle at rest |
| v_min_dps | minimum speed over the central part of the stroke |
| I_brk_A | breakaway peak current at motion start |
| I_early_A | mean current over the early stroke segment |
| I_mid_A | mean current over the middle stroke segment |
| I_late_A | mean current over the late stroke segment |
| I_latch_A | peak current during latch engagement |
| I_spike_A | largest current excursion above its moving average |
| I_rip_mA | RMS of the detrended current ripple |
| V_mean_V | mean motor voltage |
| kE_est | back-EMF constant estimated from voltage, current, and speed |
| latch_dly_s | delay from near-full stroke to switch closure |
| sw_trans | switch transition count, one for a clean closure |
| shock_g | latch impact peak |
| rebound_deg | post-latch rebound angle |
| vib_g | RMS vibration during motion |

**Table B2.** The eight health classes and their typical telemetry signatures.

| Class | Typical signature |
|---|---|
| normal | only small run-in and temperature-driven changes relative to the baseline |
| lubricant_degradation | resistance rises roughly uniformly along the stroke, all current segments and the breakaway current rise, more in the cold, and kE_est stays unchanged |
| drive_degradation | the motor torque constant drops, currents scale up multiplicatively, and kE_est falls below the design value and the baseline |
| latch_degradation | I_latch_A and latch_dly_s rise and rebound grows while shock may fall, and the rest of the stroke is unchanged |
| debris_obstruction | localized resistance mid-stroke, I_spike_A rises, v_min_dps dips, and mid-stroke current rises |
| bearing_wear | periodic torque ripple, I_rip_mA and vib_g rise strongly, shock and rebound grow, and friction rises mildly |
| harness_stiffening | resistance grows with angle, I_late_A rises much more than I_early_A, and the effect is worse in the cold |
| sensor_fault | telemetry inconsistent with physics while the mechanism is healthy, such as angle offset or gain error, switch bounce, a stuck-open switch, or current-sensor bias and noise |

**Table B3.** The four classification labels and the regression target.

| Label | Classes | Definition |
|---|---|---|
| fault_mode | 8 | One of the health classes in Table B2 |
| severity | 4 | Levels from none to severe. Severity is none for normal units. For mechanism faults it grades the loss of torque margin, and for sensor faults the size of the sensor anomaly |
| rul_bin | 4 | Bins of the cycles remaining until the torque margin turns negative under the current degradation trend |
| action | 5 | continue_nominal_operation, increase_monitoring, schedule_maintenance, stop_and_troubleshoot, verify_and_recalibrate_sensor |
| torque_margin | regression | Torque margin of the unit at the -20 C reference case, not analyzed in this paper |

# Appendix C. Full Baseline Results

Table C1 lists the 16 tabular baselines, and Table C2 lists the sequence models. XGBoost gives the best in-distribution result in both settings (0.904). Six of the sixteen tabular models stay below 0.10 on ood_unit in both settings. The tuned GRU is the best sequence model on test_id (0.803). The tuned LSTM and both RNN settings stay at or below 0.215 on every split. We treat them as failed training runs rather than as evidence about these architectures.

**Table C1.** Sixteen tabular baselines with default and Optuna-tuned hyperparameters. Each cell is the mean over 3 seeds.

| | Default hyperparameters | | | | Optuna-tuned hyperparameters | | | |
|---|---|---|---|---|---|---|---|---|
| **Model** | **test_id** | **ood_temp** | **ood_unit** | **ood_noise** | **tuned test_id** | **tuned ood_temp** | **tuned ood_unit** | **tuned ood_noise** |
| xgboost | 0.904 | 0.836 | 0.787 | 0.642 | 0.904 | 0.836 | 0.773 | 0.681 |
| histgb | 0.899 | 0.840 | 0.766 | 0.594 | 0.901 | 0.834 | 0.745 | 0.578 |
| lightgbm | 0.897 | 0.832 | 0.629 | 0.543 | 0.902 | 0.832 | 0.628 | 0.552 |
| mlp | 0.893 | 0.778 | 0.080 | 0.392 | 0.893 | 0.781 | 0.080 | 0.390 |
| catboost | 0.892 | 0.823 | 0.750 | 0.665 | 0.896 | 0.825 | 0.757 | 0.670 |
| gbdt | 0.881 | 0.772 | 0.517 | 0.687 | 0.892 | 0.811 | 0.618 | 0.649 |
| rf | 0.871 | 0.797 | 0.707 | 0.638 | 0.871 | 0.792 | 0.707 | 0.640 |
| svm_rbf | 0.863 | 0.759 | 0.081 | 0.074 | 0.884 | 0.810 | 0.081 | 0.110 |
| extratrees | 0.861 | 0.788 | 0.592 | 0.630 | 0.856 | 0.783 | 0.607 | 0.631 |
| logreg | 0.848 | 0.820 | 0.056 | 0.466 | 0.852 | 0.819 | 0.056 | 0.469 |
| dt | 0.812 | 0.727 | 0.661 | 0.532 | 0.824 | 0.739 | 0.692 | 0.511 |
| svm_lin | 0.782 | 0.746 | 0.056 | 0.475 | 0.788 | 0.737 | 0.058 | 0.472 |
| knn | 0.667 | 0.615 | 0.094 | 0.398 | 0.684 | 0.616 | 0.275 | 0.418 |
| gnb | 0.649 | 0.656 | 0.089 | 0.313 | 0.649 | 0.654 | 0.089 | 0.339 |
| lda | 0.631 | 0.623 | 0.084 | 0.520 | 0.638 | 0.619 | 0.063 | 0.500 |
| adaboost | 0.613 | 0.542 | 0.348 | 0.491 | 0.706 | 0.606 | 0.420 | 0.514 |

**Table C2.** Sequence models on the raw cycle arrays. Each cell is the mean and standard deviation over 3 seeds.

| **Model** | **Setting** | **test_id** | **ood_temp** | **ood_unit** | **ood_noise** |
|---|---|---|---|---|---|
| GRU | tuned | 0.803 ± 0.021 | 0.746 ± 0.026 | 0.627 ± 0.068 | 0.745 ± 0.040 |
| Transformer | default | 0.770 ± 0.008 | 0.738 ± 0.029 | 0.506 ± 0.086 | 0.737 ± 0.012 |
| CNN | default | 0.661 ± 0.044 | 0.603 ± 0.050 | 0.570 ± 0.077 | 0.627 ± 0.045 |
| GRU | default | 0.551 ± 0.060 | 0.483 ± 0.065 | 0.462 ± 0.063 | 0.532 ± 0.068 |
| LSTM | tuned | 0.188 ± 0.043 | 0.195 ± 0.045 | 0.215 ± 0.046 | 0.184 ± 0.044 |
| RNN | tuned | 0.172 ± 0.037 | 0.167 ± 0.032 | 0.145 ± 0.003 | 0.163 ± 0.045 |
| RNN | default | 0.169 ± 0.039 | 0.166 ± 0.039 | 0.131 ± 0.015 | 0.175 ± 0.052 |

# Appendix D. Implementation, Cost, and Budget Details

- **Student training.** Each student is fine-tuned with LoRA in bf16 with gradient checkpointing. The LoRA rank is 32, with alpha 64 and dropout 0.05. The learning rate is 2e-4, and training runs for 2 epochs with a micro batch of 8 and gradient accumulation of 2. All distilled students share these settings and the base model.

- **Baselines.** Optuna runs 30 trials per tabular model, maximizes validation macro F1, and selects hyperparameters on the validation split only. The tabular models read the same telemetry statistics that the text prompt serializes. The three recurrent architectures also run a rescue grid over learning rate and hidden width, so one failed default configuration does not end the comparison for that family.
- **Teacher access.** All teachers run through the SiliconFlow API with the same prompts (Table D1). For the probe, eight teachers ran with thinking disabled. T8 and T11 ran in default mode, because disabling thinking breaks their output format or is not supported.
- **Cost and budget.** Every API call logs its token usage, and costs come from real usage. Teacher generation cost about 115 CNY in total, and the zero-shot probe cost 7.62 CNY. A per-teacher cap of 3.00 CNY stops one teacher early with a balanced completed prefix. A global cap of 9.00 CNY stops the whole probe once the total crosses it. Generation for T5, T7, and T9 stopped early when the API budget ran out, so their retention counts only completed calls.
- **Filter diagnostics.** Drop reasons differ by teacher. Most T8 drops are parse failures (507 of 622), and almost all T11 drops are faithfulness failures (194 of 205).
- **Statistical tests.** Spearman correlations over 9 or 11 teachers use exact permutation tests, except for the rationale count, which uses a Monte Carlo permutation test. Paired bootstraps resample the test rows 10000 times and recompute macro F1 for both runs. The two-sided p-value doubles the share of resamples whose gap has the opposite sign.

Table D1 maps the teacher labels to model names, providers, and the SiliconFlow model identifiers used in our API calls.

**Table D1.** Teacher models, providers, and SiliconFlow API identifiers.

| Teacher | Model | Provider | API identifier |
|---|---|---|---|
| T1 | DeepSeek-V4.1-Flash | DeepSeek | deepseek-ai/DeepSeek-V4.1-Flash |
| T2 | DeepSeek-V3 | DeepSeek | deepseek-ai/DeepSeek-V3 |
| T3 | Qwen2.5-72B-Instruct | Qwen team, Alibaba | Qwen/Qwen2.5-72B-Instruct |
| T4 | Qwen3-32B | Qwen team, Alibaba | Qwen/Qwen3-32B |
| T5 | DeepSeek-R1 | DeepSeek | Pro/deepseek-ai/DeepSeek-R1 |
| T6 | Seed-OSS-36B-Instruct | ByteDance Seed | ByteDance-Seed/Seed-OSS-36B-Instruct |
| T7 | Qwen3.5-27B | Qwen team, Alibaba | Qwen/Qwen3.5-27B |
| T8 | Ling-flash-2.0 | inclusionAI | inclusionAI/Ling-flash-2.0 |
| T9 | Kimi-K2.6 | Moonshot AI | Pro/moonshotai/Kimi-K2.6 |
| T10 | Hy4-preview | Tencent | tencent/Hy4-preview |
| T11 | GLM-5.3 | Zhipu AI | zai-org/GLM-5.3 |